\documentclass[letterpaper, conference, 10pt]{fmt/IEEEconf}
\IEEEoverridecommandlockouts
\usepackage{cite}
\usepackage{amsmath,amssymb,amsfonts}
\usepackage{algorithmic}
\usepackage{graphicx}
\usepackage{textcomp}
\usepackage{xcolor}
\usepackage{soul}
\usepackage{comment}
\usepackage{bm}
\usepackage{multirow}

\usepackage{ulem}
\usepackage{optidef}
\usepackage{textcomp}

\usepackage{subfigure}
\usepackage{booktabs}
\usepackage{diagbox}

\newtheorem{Def}{Definition}

\usepackage{titlesec}
\titleformat{\subsubsection}[runin]{\bfseries}{\thesubsubsection}{1em}{}[: \ ]

\newcommand{\etal}{\textit{et al}. }

\DeclareMathOperator*{\argmin}{arg\,min}
\DeclareMathOperator{\atan2}{atan2}
\DeclareMathOperator{\asin}{asin}
\DeclareMathOperator{\acos}{acos}

\begin{document}

\title{\Large \bf Toward Precise Robotic Weed Flaming Using a Mobile Manipulator with a Flamethrower}

\author{Di Wang, Chengsong Hu, Shuangyu Xie, Joe Johnson, Hojun Ji, Yingtao Jiang, \\ Muthukumar~Bagavathiannan, and Dezhen Song
    \thanks{
    D. Wang, S. Xie, Y. Jiang, and D. Song are with Department of Computer Science and Engineering, Texas A\&M University. D. Song is also with Department of Robotics, Mohamed Bin Zayed University of Artificial Intelligence (MBZUAI) in Abu Dhabi, UAE. C. Hu, J. Johnson, and M. Bagavathiannan are with Department of Soil and Crop Sciences, Texas A\&M University. H. Ji is with Boston Dynamics.  Corresponding author: Dezhen Song. Email: \texttt{dezhen.song@mbzuai.ac.ae}.
    }
}
\maketitle

\begin{abstract}
Robotic weed flaming is a new and environmentally friendly approach to weed removal in the agricultural field. Using a mobile manipulator equipped with a flamethrower, we design a new system and algorithm to enable effective weed flaming, which requires robotic manipulation with a soft and deformable end effector, as the thermal coverage of the flame is affected by dynamic or unknown environmental factors such as gravity, wind, atmospheric pressure, fuel tank pressure, and pose of the nozzle. System development includes overall design, hardware integration, and software pipeline. To enable precise weed removal, the greatest challenge is to detect and predict dynamic flame coverage in real time before motion planning, which is quite different from a conventional rigid gripper in grasping or a spray gun in painting. Based on the images from two onboard infrared cameras and the pose information of the flamethrower nozzle on a mobile manipulator, we propose a new dynamic flame coverage model. The flame model uses a center-arc curve with a Gaussian cross-section model to describe the flame coverage in real time.  
The experiments have demonstrated the working system and shown that our model and algorithm can achieve a mean average precision (mAP) of more than 76\% in the reprojected images during online prediction.
\end{abstract}



\section{Introduction}

\begin{figure}[t!]
    \centering
    \includegraphics[width=3 in]{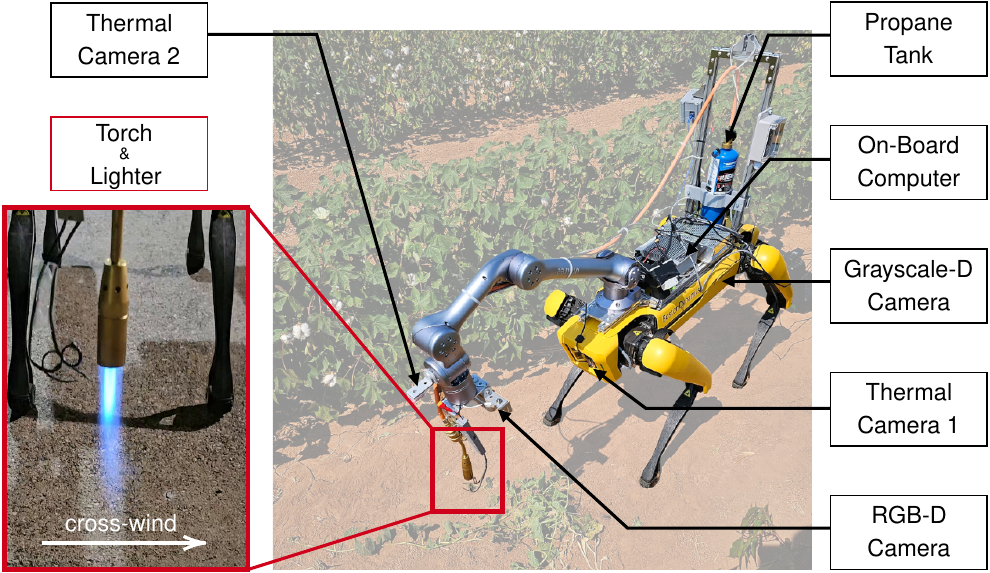}
    \caption{Robotic weed flaming system design and main components.}
    \label{fig:hardware_design}
\end{figure}

Weed infestation is an important and perpetual problem in agriculture. Weeds compete with crops for nutrients, water, and sunlight. Manual removal of weeds is labor-intensive, and large-scale mechanized herbicide spray has significant adverse impacts on the environment. Robotic removal of weeds provides a new approach to this old problem. Here, we are interested in developing environmentally friendly solutions to control weeds in early growth stages. More precisely, we report our progress in developing a robotic weed-flaming solution using a mobile manipulator equipped with a flamethrower (see Fig.~\ref{fig:hardware_design}).   

To enable precise robotic weed flaming, new system and algorithm developments are necessary.  System development includes overall design, hardware integration, and software pipeline. On the algorithm side, to enable precise weed removal, the greatest challenge is to detect and predict dynamic flame coverage in real time before motion planning. This is quite different from a conventional rigid gripper in grasping or a spray gun in painting because the thermal coverage of the flame is affected by dynamic or unknown environmental factors such as gravity, wind, atmospheric pressure, fuel tank pressure, and pose of the nozzle. Using the images from two onboard infrared cameras and the pose information of the flamethrower nozzle on a mobile manipulator, we propose a new dynamic flame coverage model. The flame model uses a center-arc curve with a Gaussian cross-section model to describe the flame coverage in real time.  

We have implemented our system and algorithm and performed physical experiments in the field. Our results show that the new design of the robotic flaming system is effective and that our flame estimation algorithm can provide a satisfactory prediction of the flame coverage. The experiments have shown that our model and algorithm can achieve a mean average precision (mAP) of more than 76\% in the re-projected images in real-time prediction. 

\section{Related Work}
This paper is closely related to robotic weed control and flame estimation. 

\subsubsection*{Robotic Weed Control}
While agricultural robotics presents new challenges to researchers \cite{Billingsley'08, Vougioukas'19}, the topics of scene perception \cite{Xie_2021,forest_perception,Agr_Stachniss_ICRA_2022,2022_weed_spray}, autonomous navigation \cite{agr_navigation-RSS-21}, motion planning \cite{Volkan_Weed_Mowing}, and manipulation \cite{Davidson'20, Vrochidou'22,Meshram'22} in agricultural environment have been extensively explored. 

For robotic weed control, existing researches have covered weed detection \cite{Xie_2021} and various control mechanisms, such as cultivation \cite{mccool_robotic_weeding_tools}, stamping \cite{michaels_stamping}, mowing \cite{melita_mow}, and precise application of herbicides \cite{lee_spray,2022_weed_spray}. As an alternative to these mechanical or chemical weed control methods, weed flaming is contact-free and has been shown to be effective on herbicide resistant weed species while capable of preserving plant residues to reduce water erosion and leaving no chemical residues \cite{bauer2020thermal}. The non-selective characteristic of weed flaming makes it applicable to various kinds of weeds, but it also poses a challenge on precise flame control to avoid collateral damage to the field. More specifically, flaming should desiccate the weed with minimal disturbance to crops or soil microorganisms \cite{Rahkonen'99, bauer2020thermal}. Studies on manual weed flaming have explored feasibility \cite{vsniauka2008thermal}, effectiveness~\cite{ascard1998comparison}, best application conditions~\cite{bauer2020thermal}, impact of burning angles, burner designs and shielding methods on weed flaming \cite{storeheier1993basic}. Due to the high labor cost and the lack of precise flame control methods, the application of manual weed flaming is limited. Meanwhile, robotic weed flaming has not been well studied, which presents unique problems for us to explore.

\subsubsection*{Flame Estimation}
Robotic weed flaming poses a unique challenge on modeling the flame as a soft end-effector in dynamic cross-wind environment, which is unprecedented in other robotic manipulation applications. Rigid end-effectors are commonly used in object grasping, crop harvesting and manufacturing \cite{Babin'21, Davidson'20}, and their behaviors can be modeled by forward kinematics \cite{Mason'01}. As for soft end-effectors, they are typically modeled based on their individual physical properties. Notable examples that use gas or liquid flows as soft end-effectors include air levitation conveyors \cite{Delettre'11, Laurent'11}, surgical water jets \cite{Morad'15} and herbicide/paint sprayers \cite{Potkonjak'00, From'11, Chen'17, 2022_weed_spray}. Similar to these soft end-effectors, the flame for weed removal needs a model that can capture its deformation caused by the outdoor environment and the fluctuation in gas fuel pressure. In this paper we propose a data driven model to estimate the flame direction and coverage without explicitly deriving the complex physical model since it is not necessary for our application.


The combustion process of flame has been studied extensively in the field of energy engineering. Modeling and estimating the flame direction and coverage for weed removal is closely related to the topics of flame tomography reconstruction and flame radiation estimation.

Flame tomography reconstruction aims to recover the flame temperature distribution using the computed tomography technique. By observing the flame emission or absorption spectrum from multiple perspectives \cite{Jin'22, Liu'18}, the flame temperature distribution is reconstructed offline as volumetric grids. Flame tomography reconstruction can achieve superior spatial and temporal resolution by taking measurements at regular intervals using camera/photodiode arrays \cite{MA'16,CAI'17}. In \cite{MA'16}, the authors reconstructed a $70\times 70\times 105 \text{ mm}^{3}$ flame volume at 5kHz using six cameras and two workstations with over 200GB of RAM. Although flame tomography reconstruction is ideal for detailed profiling of the indoor flaming process, its complicated setup and computational cost make it impractical in field applications.

Flame radiation estimation focuses on predicting the transfer of flame energy in outdoor environments. Semi-empirical models that capture flame characteristics, such as width and center line length, are widely adopted to avoid the complexity of establishing an accurate physical model of outdoor flame radiation \cite{Aziz'19}. Wang \etal \cite{Wang'22} proposed a 2D flame length prediction model for propane jet fires in a crosswind environment based on the circular arc flame center line assumption. Zhou \etal \cite{Zhou'15} presented a line source radiation model and measured three different flame shapes to predict the flame heat flux profile. These semi-empirical models provide valuable insights of the flame characteristics in outdoor environments, but their focus on large scale flame width and center line estimation in 2D makes them unsuitable for 3D flame coverage estimation in fields.

Building on the existing methods, we are developing a 3D flame surface model that requires measurements from as little as two camera perspectives and can be used in real-time field applications.




\section{System Design}

We begin with the overall design of the hardware system before introducing the software diagram.

\subsection{Hardware System}


The system uses the Spot Mini\texttrademark\  quadruped robot from Boston Dynamics\texttrademark\  as the moving platform with a light weight 6-DoF Unitree\texttrademark\ Z1 manipulator/arm mounted on the back. The arm has a payload capacity of 2kg. The Spot and the arm are both powered by the Spot's internal battery, which provides 58.8V direct current (DC). A step-down DC-DC converter is used to provide power for the arm and on-board computer. An Intel Realsense\texttrademark\  D435 RGB-D camera is mounted on the manipulator end-effector for weed detection and localization. Two FLIR\texttrademark\ Lepton 3.5 thermal cameras are mounted on the end-effector and robot body to monitor the flame direction and coverage. The arm, thermal cameras and RGB-D camera form a hand-eye vision system, as this configuration allows the system to obtain observations of the weed scene and torch flame at different angles and distances. All perception, decision-making, and control algorithms are executed on an on-board computer which is the Spot Core from Boston Dynamics\texttrademark\ . To provide the flaming capability, a propane tank is mounted on the Spot, and the propane gas is delivered to the torch mounted at the tip of the arm through a torch control unit. The unit receives control commands from the onboard computer and turns on/off the gas flow when the torch is at the desired actuation position. A relay-controlled lighter mounted next to the torch is used to ignite the fuel.

\begin{figure}[htbp]
    \centering
    \includegraphics[height=1.8in]{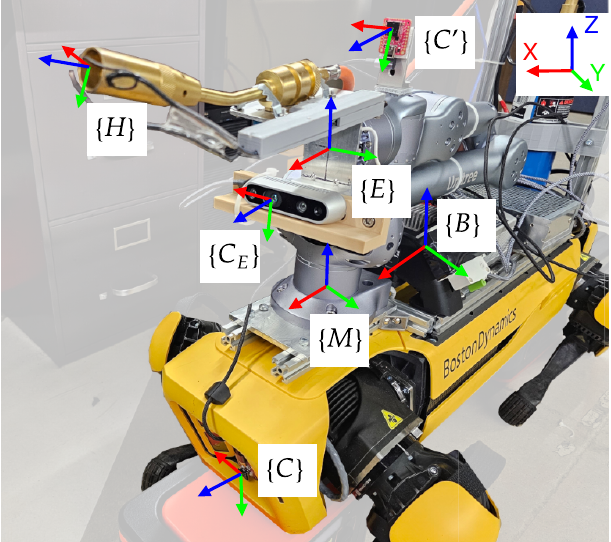}
    \caption{Illustration of the robot coordinate systems with color-coded axes.}\label{fig:coordinate-system}
\end{figure}

The coordinate systems of our weed removal robot are shown in Fig.~\ref{fig:coordinate-system}. The two thermal camera frames $\{C\}$, $\{C'\}$ and the frame of the RGB-D camera which is mounted on the end-effector $\{C_E\}$ are calibrated to the manipulator frame $\{M\}$ and the end-effector frame $\{E\}$ using hand-eye calibration. The transformation between the torch nozzle head $\{H\}$ and the cameras $\{C'\}$ and $\{C_E\}$ is established using the extrinsic calibration of the camera with a custom designed checkerboard mounted on the nozzle head. The Spot robot body frame $\{B\}$ and the manipulator frame $\{M\}$ are calibrated using the hand-eye calibration results and extrinsic camera calibration between the end effector RGB-D camera and the robot Grayscale-D camera.

\begin{figure}[htbp]
    \centering
    \includegraphics[width=3.2in]{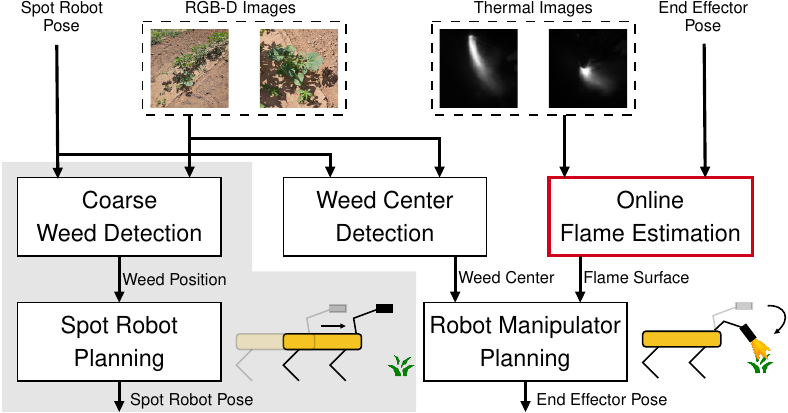}
    \caption{Software diagram. The software system is partitioned into two components with the shaded blocks control Spot body motion and the rest directs the manipulator motion for weed flaming.}   \label{fig:software_diagram}
\end{figure}
\subsection{Software Design}

Fig.~\ref{fig:software_diagram} illustrates the software pipeline for our robotic weed flaming system. The data processing procedure begins with a preliminary coarse localization of the weed, which uses the weed detection model trained within the YOLOv6 framework\cite{li2022yolov6}. Subsequently, the Spot robot planning component generates a trajectory, using the Spot software development kit (SDK), thus guiding the Spot robot from its initial state to a position in close proximity to the identified weed. This step ensures that the weed is within the region that allows the manipulator to perform the flaming operation (the gray box in Fig.~\ref{fig:software_diagram}). More details on the planning algorithm can be found in \cite{xie2023coupled}. After Spot arrives at the planned position, we perform weed center detection using the close-up view from the RGB-D camera. Meanwhile, the online flame estimation is performed to estimate the temperature thresholded flame surface for precise manipulation. As the last step, the manipulation planning finds the state for the flaming pose of the end effector using the combined position of the center of the weed and the flame surface. The motion plan of the arm is computed using the Unitree\texttrademark\ arm SDK.

The system is an integration of many existing developments with an important algorithmic challenge which is online flame estimation. Unlike fixed end-effectors in other robotic manipulation tasks, the fire-flame front is not a rigid body, and understanding its coverage in real time determines the success of the weed-flaming task. In the following, we will explain how we address this problem. 

\begin{figure}[!tbp]
    \centering
    \subfigure[]{\includegraphics[width=1.5in]{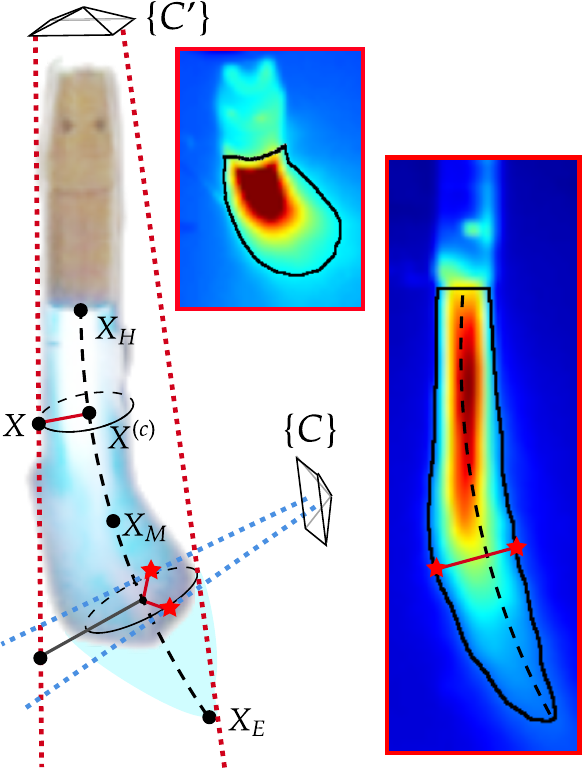}\label{fig:flame-projection}}
    \hspace*{.2in}
    \begin{minipage}{1in}
    \vspace*{-1.4in}
    \subfigure[]{\includegraphics[width=1.1in]{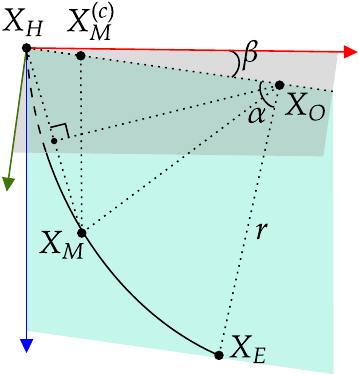}\label{fig:flame-model}}\\
    \subfigure[]{\includegraphics[width=1.1in]{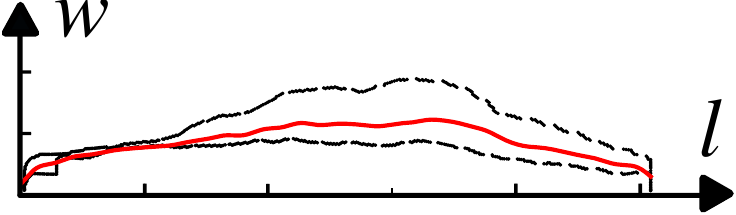}\label{fig:flame-width}} 
    \end{minipage}
    \caption{(a) Illustration of flame projection. The projected thermal images in the first and second views are shown in the short and long red boxes, respectively. Using the center line (black dashed line), each backprojected ray from $\{C\}$ (blue dotted lines) determines a flame surface point (red stars), while the depth ambiguity in rays from $\{C'\}$ (red dotted lines) blocks the surface point estimation. (b) Illustration of the flame center arc model. (c) Illustration of the flame cross-section width function $w(l)$. The red line shows the fitted $w(l)$ and the black dots are the measurements}\label{fig:electronic-components-signals}
\end{figure}
\section{Problem Formulation}

The online flame coverage estimation problem takes images from two thermal cameras to estimate flame coverage. As shown in Fig.~\ref{fig:hardware_design}, thermal camera 1 is installed in front of the Spot body and provides a side view of the flame. Thermal camera 2 is installed on the end effector of the arm. 
\subsection{Assumption}
To allow us to focus on the key problem, we have the following assumptions.  
\begin{itemize}
\item[a.1] The flame cross-section temperature distribution is a 2D isotropic Gaussian distribution.
\item[a.2] The initial flame direction is the same as the torch direction.
\item[a.3] The flame radiates uniformly over its boundary surface \cite{MUDAN'84, Luketa'22}.
\item[a.4] All cameras and the manipulator have been calibrated and hence the intrinsics of the cameras and relative poses among frames in Fig.~\ref{fig:coordinate-system} are known.
\item[a.5] The ambient thermal radiation is constant. 
\end{itemize}


\subsection{Nomenclature}
\begin{itemize}
\item[$\{H\}$] is the 3D frame of the torch nozzle head. Its origin is at the center of the nozzle head. Its Z-axis is parallel to the direction of the nozzle. All variables are defaulted to $\{H\}$.
\item[$\{C\}$] is the 3D frame of the first thermal camera. Its origin is at the camera optical center, its $Z$-axis coincides with the optical axis and pointing to the forward direction of the camera. Similarly, we define $\{C'\}$ as the 3D frame for the second thermal camera.
\item[$\mathbf{\tilde{x}}, \mathbf{\tilde{x}}'$] are points in the first and second views, $\mathbf{\tilde{x}}, \mathbf{\tilde{x}}' \in \mathbb{R}^2$, respectively. The homogeneous counterparts of $\mathbf{\tilde{x}}$ is $\mathbf{x} := [\mathbf{\tilde{x}}^\mathsf{T} \textrm{  }  1]^\mathsf{T}  \in \mathbb{P}^2$, and $\mathbf{x}':= [\mathbf{\tilde{x}}^\mathsf{\prime T} \textrm{  }  1]^\mathsf{T}  \in \mathbb{P}^2$.
\item[$\mathrm{\tilde{X}}$] is a point in the 3D Euclidean space, $\mathrm{\tilde{X}} \in \mathbb{R}^3$. Its homogeneous counterpart is $\mathrm{X} = [\mathrm{\tilde{X}}^\mathsf{T} \textrm{  }  1]^\mathsf{T} \in \mathbb{P}^3$.
\item[$F(\mathrm{X})$] is the flame surface function, $F(\mathrm{X})=0$ if and only if $\mathrm{X}$ is on the surface.
\end{itemize}
By convention, the prime symbol $'$ denotes variables in the second thermal camera view. Let us define our problem.
 
\subsection{Problem Definition}
\begin{Def}[Flame Estimation]\label{def:flame_est}
Given a temperature threshold $T$, thermal images $I$ and $I'$ from two different perspectives, and the two-view camera poses ${}_{H}^{C}\mathbf{T}$ and ${}_{H}^{C'}\mathbf{T}$, estimate the temperature thresholded flame boundray surface function $F(\mathrm{X})$.
\end{Def}

\section{Algorithm}\label{sec:algorithm}

\subsection{Flame Modeling}\label{sec:flame-modeling}
We model the flame with a center curve associated with a circular cross-section based on the 2D isotropic Gaussian distribution cross section temperature assumption. 
Let $f_c$ be the flame center curve function such that $\mathcal{C} = \{\mathrm{X}_\mathcal{C} | f_c(\mathrm{X}_\mathcal{C})=0\}$ are all the points on the curve. The flame cross-section at point $\mathrm{X}_\mathcal{C}$ is a circle centered at $\mathrm{X}_\mathcal{C}$ and perpendicular to the flame center curve. For an arbitrary point $\mathrm{X}$, its closest corresponding point on the center curve is $\mathrm{X}^{(c)} = \displaystyle \argmin_{\mathrm{X}_i\in \mathcal{C}} \| \mathrm{X}-\mathrm{X}_i\|_2$, where $\|\cdot\|_2$ is the $\mathrm{L}_2$ norm. Based on this flame center curve $f_c$, the temperature
thresholded flame surface function is modeled as
\begin{equation}\label{eq:FlameSurface}
F(\mathrm{X}) = \|\mathrm{X}-\mathrm{X}^{(c)}\|_2 - w(l_{\mathrm{X}^{(c)}})
\end{equation}
where $w(l_{\mathrm{X}^{(c)}})$ is the frame cross-section width function that encodes the radius of the cross-section boundary circle centered at $\mathrm{X}^{(c)}$, and $l_{\mathrm{X}^{(c)}}$ is the curve length from nozzle position $\mathrm{X}_H$ to $\mathrm{X}^{(c)}$. As shown in Fig.~\ref{fig:flame-projection}, the distance between $\mathrm{X}$ and the center curve is equal to the cross section width (red line) only when $\mathrm{X}$ is a point on the cross-section boundary circle. Therefore $F(\mathrm{X})=0$ defines the boundary surface of the flame.

To model the flame center curve, we extend the approximation proposed in \cite{Wang'22} and use a 3D circular arc representation. As shown in Fig. \ref{fig:flame-model}, because the flame starts from the torch nozzle and its initial direction is the same as the torch direction, the tangent direction of the circular arc at the position of the nozzle $\mathrm{X}_H$ is parallel to the Z axis of the torch frame, which means that the center of the circular arc $\mathrm{X}_O$ must lie in the X-Y plane of the torch frame. Intuitively, this circular arc can be represented by its radius $r$, central angle $\alpha$ and rotation angle $\beta$ along Z axis of the torch frame. Here we use the circular arc midpoint $\mathrm{X}_M$ as an alternative representation for simplicity. Because $\mathrm{X}_M$ is a point on the arc, line $\overline{\mathrm{X}_{H}\mathrm{X}_{M}^{(c)}}$ formed by $\mathrm{X}_H$ and the X-Y plane projection of $\mathrm{X}_M$ must pass through $\mathrm{X}_O$. The intersection of the perpendicular bisector of $\overline{\mathrm{X}_{H}\mathrm{X}_{M}}$ and $\overline{\mathrm{X}_{H}\mathrm{X}_{M}^{(c)}}$ is the arc center $\mathrm{X}_O$. While deriving $\mathrm{X}_O$ from $\mathrm{X}_{M}$ resolves radius $r$ and rotation angle $\beta$, central angle $\alpha$ can be recovered from the fact that $\angle \mathrm{X}_H\mathrm{X}_M\mathrm{X}^{(c)}_M$ is equal to $\frac{\alpha}{4}$. Therefore, the flame center circular arc parameters can be fully represented by the circular arc midpoint $\mathrm{X}_M$. We have $\alpha=4\asin\frac{\|[\mathrm{X}_M]_{1:2}\|_2}{\|\mathrm{X}_M\|_2}$, $\beta = \atan2([\mathrm{X}_M]_2, [\mathrm{X}_M]_1)$ and $r = \frac{\|\mathrm{X}_M\|_2}{2}/\sin\frac{\alpha}{4}$, where $[\mathrm{X}_M]_{1:2}$ is the first two components of $\mathrm{X}_M$. Using the midpoint representation $\mathrm{X}_M$ also enhances the robustness during parameter estimation. When the flame center curve approaches a straight line, the estimated midpoint will approach the Z axis of the torch frame, which avoids the numerical instability caused by the circular arc radius approaching infinity.

In the following part of this section, we will show the estimation process of the flame center circular arc and the cross-section width function.

\subsection{Flame Center Curve Estimation}
Because the flame center arc parameter $\mathrm{X}_M$ is estimated from points in the thermal images, let us first introduce the camera projection model. Based on the pinhole camera model \cite{hartley'04}, a 3D point $\mathrm{X}$ and its counterpart $\mathbf{x}$ in the camera image satisfy the following.
\begin{equation} \label{eq:CameraModel}
\mathbf{x} = \lambda\mathbf{K}[\mathbf{R} \textrm{  } \mathbf{t}]\mathrm{X}.
\end{equation}
Here $\lambda$ is a scaling factor, $\mathbf{K}$ is the intrinsic matrix of the camera, $\mathbf{R}$ and $\mathbf{t}$ are the components of rotation and translation of the transformation matrix ${}_{H}^{C}\mathbf{T} = \begin{bmatrix}\mathbf{R} & \mathbf{t} \\ \mathbf{0} & 1\end{bmatrix}$, which transform from frame $\{H\}$ to frame $\{C\}$. ${}_{H}^{C}\mathbf{T}$ is obtained from the forward kinematics of the manipulator and the extrinsic calibration of the thermal cameras and the torch. Similarly, we can define $\mathbf{x}'$, $\lambda'$, $\mathbf{K}'$, $\mathbf{R}'$, and $\mathbf{t}'$ for the second view, which follow the same pinhole model described in \eqref{eq:CameraModel}.

By color thresholding the thermal image from the first view camera, we extract the center point $\mathbf{x}_c$ of the thresholded area. This center point is a projection of the circular arc midpoint $\mathrm{X}_M$, therefore they satisfy
\begin{equation} \label{eq:Midpoint}
[\mathbf{x}_c]_{\times}\mathbf{K}[\mathbf{R} \textrm{  } \mathbf{t}]\mathrm{X}_M=\mathbf{0}.
\end{equation}
Here $[\cdot]_{\times}$ is the skew-symmetric matrix. For the second view, \eqref{eq:Midpoint} can be rewritten by replacing $\mathbf{x}_c$, $\mathbf{K}$, $\mathbf{R}$ and $\mathbf{t}$ with their counterparts. The flame center arc midpoint $\mathrm{X}_M$ is solved from the two view stacked version of \eqref{eq:Midpoint}, $\mathbf{A}\mathrm{X}_M=\mathbf{0}$, where $\mathbf{A} = \begin{bmatrix} [\mathbf{x}_c]_{\times}\mathbf{K}[\mathbf{R} \textrm{  } \mathbf{t}] \\ [\mathbf{x}'_c]_{\times}\mathbf{K}'[\mathbf{R}' \textrm{  } \mathbf{t}']\end{bmatrix}$.

Based on the estimate midpoint $\mathrm{X}_M$, the flame center curve point set $\mathcal{C}$ is assembled with points $\mathrm{X}_\mathcal{C}$ derived from the estimated midpoint $\mathrm{X}_M$ as
\begin{equation} \label{eq:CenterCurve}
\mathrm{X}_\mathcal{C} = 
      \begin{bmatrix} \mathbf{R}_Z(\beta) & \mathrm{X}_O \\ \mathbf{0} & 1\end{bmatrix} 
      \begin{bmatrix} r\cos a \\ 0 \\ r\sin a \\ 1\end{bmatrix}, 
      \text{\quad} a\in[0,\alpha].
\end{equation}
where $\mathbf{R}_Z$ returns the Z-axis rotation matrix, $\mathrm{X}_O = \mathbf{R}_Z(\beta)[r \ 0 \ 0]^{\mathsf{T}}$, $\alpha$, $\beta$ and $r$ are circle arc parameters derived in Sec.~\ref{sec:flame-modeling}. 

\subsection{Flame Cross-Section Width Estimation}
Similar to the classical 3D shape-from-silhouette 3D volumetric reconstruction techniques \cite{Martin'83, Kutulakos'97}, we use the thresholded boundary points in image to first estimate the 3D flame surface points, then calculate the width of their corresponding cross-sections and use Gaussian process (GP) regression to model the continuous cross-section width function.


\subsubsection*{Flame Surface Point Estimation}
We recover the flame surface points based on thresholded boundary points in two views and the estimated flame center curve.


If a flame surface point projects to thresholded boundary points in both views, it can be recovered from the back-projected rays of these boundary points. However, due to the pixelization error of cameras, the back-projected rays might not intersect with each other. We set a distance threshold $d$ to resolve this issue, we register the closest point to both rays as a surface point when the distance between the two rays is less than $d$, which is analogous to the voxel size in the volumetric methods. The distance between rays is also used to identify the correspondences between the boundary points in two views by iterating through all their combinations in the order of their distance to the epipolar line. Let $\mathbf{x}_{j}$ be the $j$-th point on the thresholded boundary in the first view image, and $\mathbf{x}'_{k}$ be the $k$-th boundary point in the second view. Based on \eqref{eq:CameraModel}, the direction vector of the back-projected ray from $\mathbf{x}_{j}$ is $\mathbf{v}_{j} = \mathbf{R}^{\mathsf{T}}\mathbf{K}^{-1}\mathbf{x}_{j}$ and the ray point set is $\mathcal{L}_{j} = \{\mathrm{X}_{j}| \mathrm{\tilde{X}}_{j} = \lambda\mathbf{v}_{j}-\mathbf{R}^{\mathsf{T}}\mathbf{t}; \lambda\in\mathbb{R}\}$. Similarly, $\mathbf{v}'_{k}$ and $\mathcal{L}'_{k}$ can be derived from $\mathbf{x}'_{k}$. Given these two back-projected rays $\mathcal{L}_{j}$ and $\mathcal{L}'_{k}$, their closest points to each other are $\mathrm{\hat{X}}_{j}, \mathrm{\hat{X}}'_{k} = \argmin_{\substack{\mathrm{X}_{j}\in \mathcal{L}_{j} \\ \mathrm{X}'_{k}\in \mathcal{L}'_{k}}} \|\mathrm{X}_{j} - \mathrm{X}'_{k}\|_2$. And their acute angle between each other is $\theta_{j,k} = \acos(|\frac{\mathbf{v}_{j}^{\mathsf{T}}\mathbf{v}'_{k}}{|\mathbf{v}_{j}||\mathbf{v}'_{k}|}|)$. A flame surface point is registered to the surface point set $\mathcal{S}$ as
\begin{equation} 
\resizebox{0.9\linewidth}{!}{
  $\mathrm{X}_s = \frac{\mathrm{\hat{X}}_{j}+\mathrm{\hat{X}}'_{k}}{2} \in \mathcal{S}, 
  {\text{~ if}\ (\|\mathrm{\hat{X}}_{j}-\mathrm{\hat{X}}'_{k}\|_2<d)\wedge 
  (\theta_{j,k} >\theta})$.
}
\end{equation}
Here $d$ identifies their correspondence, and $\theta$ determines the acceptable uncertainty range of this estimation, a smaller $\theta$ means the acceptable uncertainty range is larger.

If a flame surface point projects to thresholded boundary points in only one view, it can be recovered from the flame center curve and the back-projected ray of this boundary point. Due to depth ambiguity, this surface point can be uniquely identified only when the back-projected ray is perpendicular to the center curve (blue dotted line in Fig.~\ref{fig:flame-projection}), otherwise the surface points on the back-projected ray are indistinguishable (points on red dotted line in Fig.~\ref{fig:flame-projection}), potentially leading to a wrong cross-section width estimation (black line in Fig.~\ref{fig:flame-projection}). Let the back-projected ray of the j-th boundary point in the first view be $\mathcal{L}_j$, its direction vector is $\mathbf{v}_j$, and the gradient direction of the center curve $\mathcal{C}$ at point $\mathrm{\hat{X}}_c$ is $\mathbf{v}_c$. The closest points between them are $\mathrm{\hat{X}}_{j}, \mathrm{\hat{X}}_{c} = \argmin_{\substack{\mathrm{X}_{c}\in \mathcal{C} \\ \mathrm{X}_{j}\in \mathcal{L}_{j}}} \|\mathrm{X}_{j} - \mathrm{X}_{c}\|_2$. A flame surface point is registered to the surface point set $\mathcal{S}$ as
\begin{equation} 
\mathrm{X}_s = \mathrm{\hat{X}}_{j}\in\mathcal{S}, \text{ \quad if } \mathbf{v}_c^{\mathsf{T}}\mathbf{v}_j = 0.
\end{equation}
For boundary points in the second view, their corresponding surface points can be derived in the same way. The perpendicular condition can be relaxed to $\acos(|\frac{\mathbf{v}_{j}^{\mathsf{T}}\mathbf{v}_{c}}{|\mathbf{v}_{j}||\mathbf{v}_{c}|}|)>\gamma$ when certain amount of depth ambiguity is acceptable.

\subsubsection*{Cross-Section Width Function Estimation}
With the estimated flame surface points $\mathrm{X}_s \in \mathcal{S}$, we first obtain their closest point $\mathrm{X}_s^{(c)}$ to the center curve in the same way as \eqref{eq:FlameSurface}, then calculate their corresponding cross-section widths $w_s = \|\mathrm{X}_s-\mathrm{X}_s^{(c)}\|_2$ and circular arc length $l_s = 2r\asin(\frac{\|\mathrm{X}^{(c)}_s-\mathrm{X}_H\|_2}{2r})$. Using the training data, we fit the cross-section width prediction function $w(l)$ using GP regression as
\begin{equation} \label{eq:WidthFunction}
\resizebox{0.9\linewidth}{!}{
$w(l) \sim \mathcal{GP}\bigl(\mathbf{k}_*^{\mathsf{T}}(\mathrm{K}+\sigma_N^2\mathbf{I})^{-1}\mathbf{w},\text{ } k_{**}-\mathbf{k}_*^{\mathsf{T}}(\mathrm{K}+\sigma_N^2\mathbf{I})^{-1}\mathbf{k}_*\bigr)$
}
\end{equation}
where $\mathbf{w} = [w_1 \ldots w_N]^{\mathsf{T}}$ are the training cross-section widths, $\sigma_N$ is the observation variance. Here $\mathrm{K}\in \mathbb{R}^{N\times N}$, $\mathbf{k}_*\in \mathbb{R}^{N\times 1}$ and $\mathrm{k}_{**}\in \mathbb{R}$ are the covariances of the training-training, training-query, and query-query points, respectively. Their individual component $\mathrm{K}[m,n] = \mbox{cov}(l_m,l_n)$, and here $\mbox{cov}(\cdot,\cdot)$ is the thin-plate function \cite{williams'07}. Fig.~\ref{fig:flame-width} shows an example of the fitted $w(l)$ and the measurements.


\subsection{Joint Optimization}
With the estimated flame center curve and cross-section width, we first evaluate the re-projected flame silhouett in two views, then refine the estimation based on the intersection over union (IoU) cost. 

Let $\mathcal{I}$ be the set of silhouette points in the first view obtained through thresholding the original thermal image, and $\mathcal{I}'$ is its counter part in the second view. The set of re-projected silhouette points in the first view is
\begin{equation} \label{eq:re-projection}
_{ } \mathcal{I}_r=\{\mathbf{x}_j \text{ } | \text{ } \exists \mathrm{X}_{j}\in\mathcal{L}_{j}\text{, }F(\mathrm{X}_{j})\leq 0\}.
\end{equation}
And let $\mathcal{I}_r'$ be its counter part in the second view. We jointly optimized the flame center arc midpoint and the estimated flame widths $\mathcal{X} = [\mathrm{X}_M^\mathsf{T} \ \mathbf{w}^\mathsf{T}]^\mathsf{T}$ using the IoU as
\begin{equation}
   \max_{\mathcal{X}} \ \frac{|\mathcal{I}\cap \mathcal{I}_r|}{|\mathcal{I}\cup \mathcal{I}_r|}+\frac{|\mathcal{I}'\cap \mathcal{I}'_r|}{|\mathcal{I}'\cup \mathcal{I}'_r|}.
\end{equation}

\section{Experiments}
We have constructed the system and tested the entire system and the online flame estimation algorithm under field conditions. First, let us show the flame estimation results.





\subsection{Flame Estimation Validation}
To validate the proposed online flame estimation algorithm, we have collected a dataset with 156 pairs of images from the thermal cameras in an indoor environment where the speed and direction of the wind can be manually controlled. We manually classified the collected images as being affected by light/strong wind and used the procedure described in Sec.~\ref{sec:algorithm} estimate the circular arc flame model. Additionally, we also estimated the flame based on a straight line model to server as a baseline. We can not compare our method with the conventional flame reconstruction methods \cite{MA'16, Jin'22} because they require images from multiple side-view cameras, which are not avaliable on the weed flaming robot. The accuracy of the estimations is evaluated by the mean average precision (mAP) of the re-projected images of the reconstructed 3D flame model with respect to the temperature thresholded thermal images.


\begin{figure}[!tbp]
    \centering
    \subfigure[]{\includegraphics[width=1in]{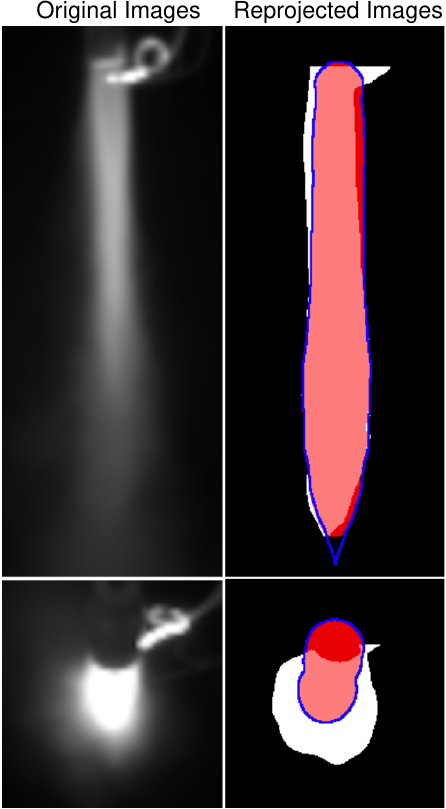}\label{fig:flame-nowind}}
    \hspace*{.1in}
    \subfigure[]{\includegraphics[width=1in]{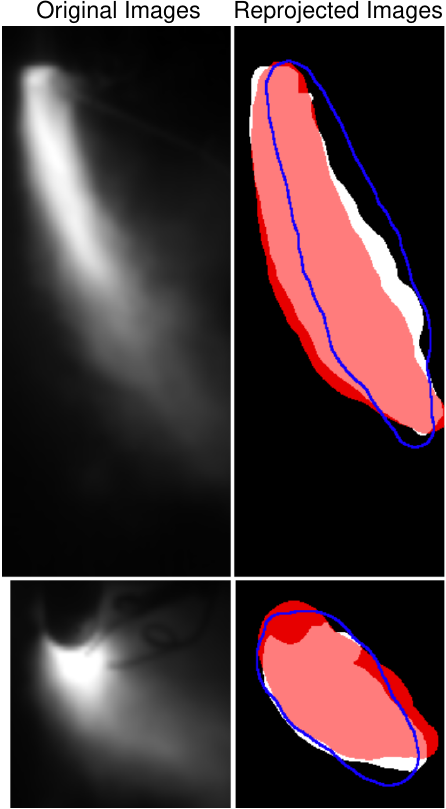}\label{fig:flame-strongwind}}
    \hspace*{.1in}
    \begin{minipage}{0.6in}
    \vspace*{-1.3in}
    \subfigure[]{\includegraphics[width=0.6in]{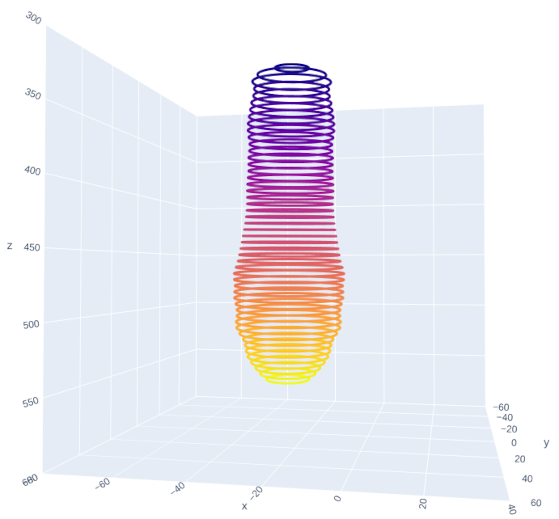}\label{fig:flame-nowind-3d}}\\
    \subfigure[]{\includegraphics[width=0.6in]{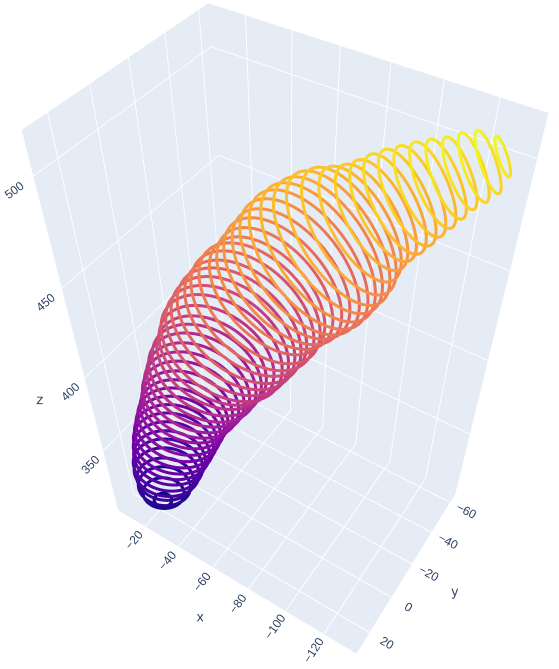}\label{fig:flame-strongwind-3d}}
    \end{minipage}
    \caption{(a) and (b) show the original thermal images and the reprojected images under light wind and strong wind, respectively. The red area is from the circular arc model, the white area is the thresholded region and the bule outline is from the straight line model. (c) and (d) show the corresponding reconstructed 3D flame using the circular arc model.}\label{fig:flame-results}
\end{figure}

\begin{table}[!htbp]
	\centering
	\caption{Flame Estimation Mean Average Precision}\label{tab:flame-est}
  \begin{tabular}{l c c c}
          \toprule[0.8pt]
           &   Light Wind    &   Strong Wind  &    Overall     \\\midrule
          Straight Line & 75.8\% & 68.6\% & 71.5\% \\
          Circular Arc & 79.7\% & 74.5\% & 76.6\% \\
          \hline
  \end{tabular}
\end{table}

\begin{figure}[!tbp]
    \centering
    \subfigure[]{\includegraphics[height = 1in]{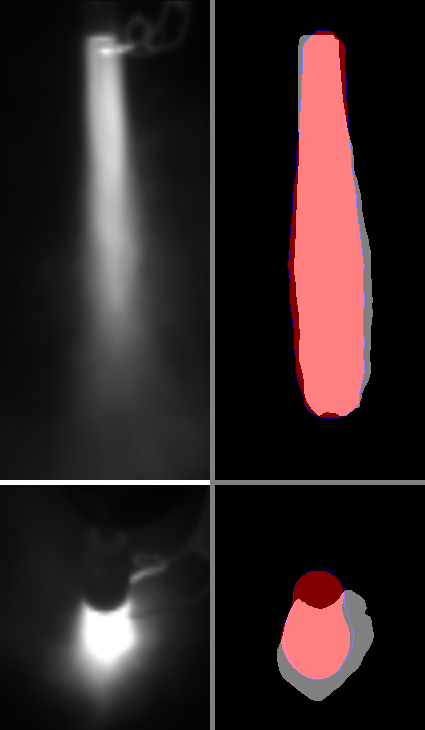}}
    \subfigure[]{\includegraphics[height = 1in]{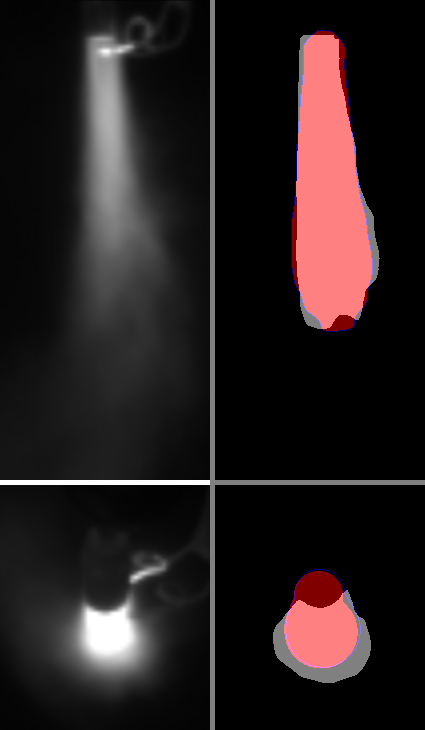}}
    \subfigure[]{\includegraphics[height = 1in]{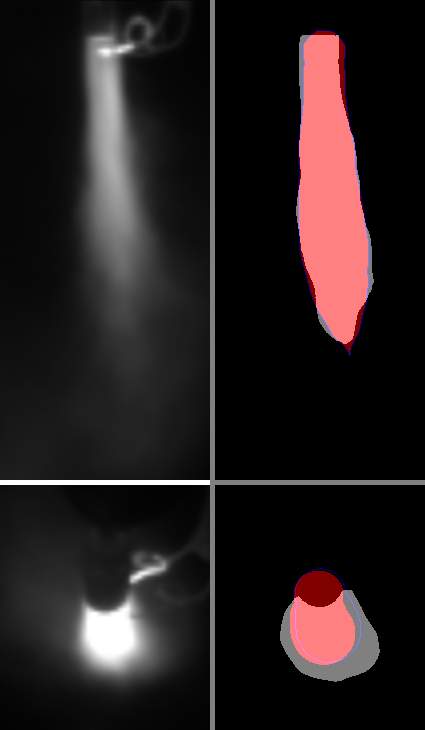}}
    \subfigure[]{\includegraphics[height = 1in]{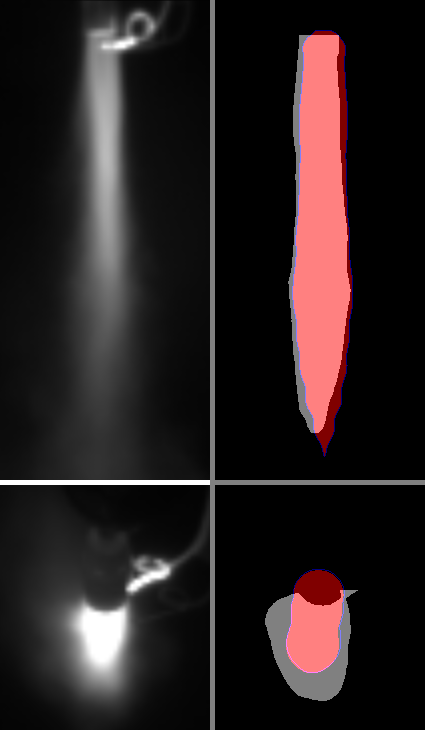}}
    \subfigure[]{\includegraphics[height = 1in]{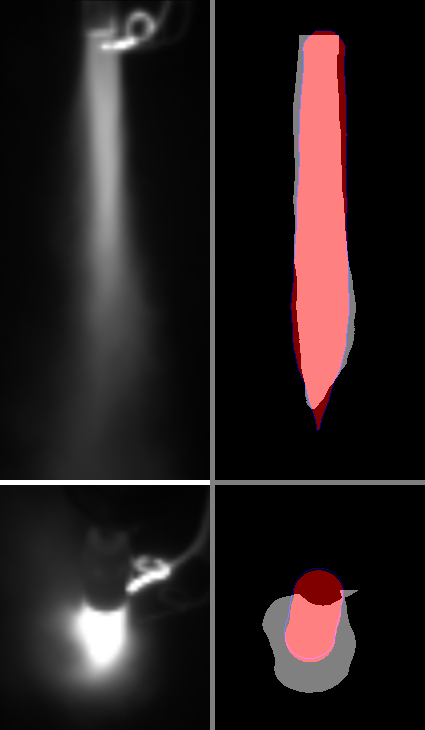}} \\
    \subfigure[]{\includegraphics[height = 1in]{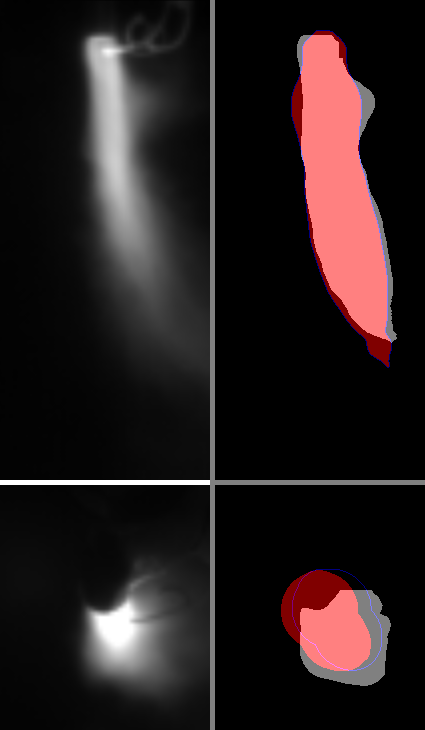}}
    \subfigure[]{\includegraphics[height = 1in]{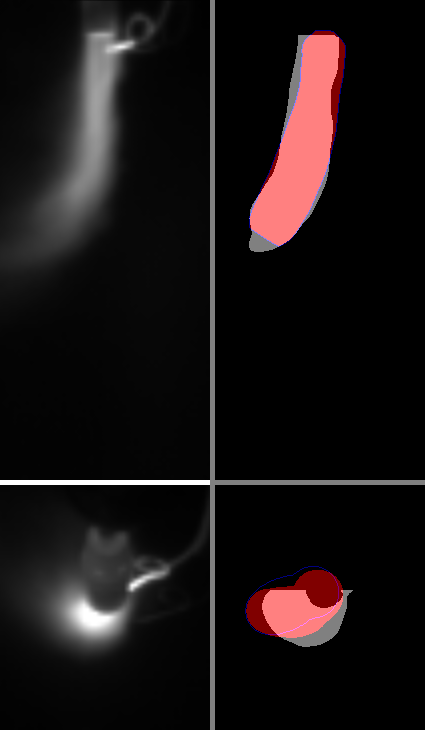}}
    \subfigure[]{\includegraphics[height = 1in]{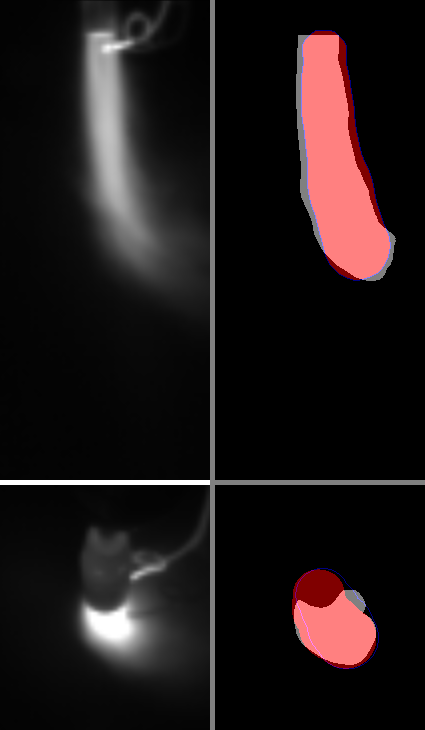}}
    \subfigure[]{\includegraphics[height = 1in]{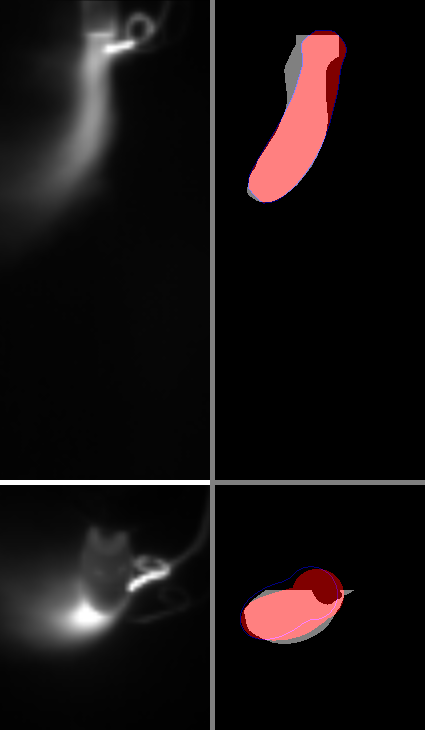}}
    \subfigure[]{\includegraphics[height = 1in]{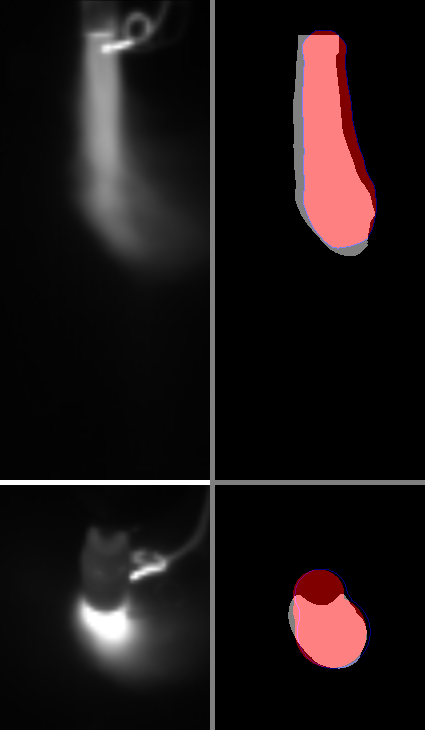}} \\
    \caption{Typical results of (a)-(e) the light wind dataset and (f)-(j) the strong wind dataset.}\label{fig:flame-results-ext}
\end{figure}

As shown in Tab.~\ref{tab:flame-est}, the circular arc model outperforms the baseline straight line model in both light wind and strong wind conditions. Fig.~\ref{fig:flame-results} shows the example original and reprojected images under light/strong wind. The circular arc model and the straight line model produce similar results under light wind (Fig.~\ref{fig:flame-nowind}), but the performance of the straight line model degrades under strong wind while the circular arc model maintains its performance (Fig.~\ref{fig:flame-strongwind}). More results from the light-wind and strong-wind datasets are shown in Fig.~\ref{fig:flame-results-ext} and the multimedia attachment.

\subsection{Weed Flaming System Validation}
We conducted physical experiments on a raised bed plot and a cotton field to validate our flaming system against different weeds. We present quantitative results of the raised bed plot experiment in this section and qualitative results of the cotton field experiment in the multimedia attachment.

The experiment in the raised-bed plot focuses on quantitatively testing the full system pipeline from weed detection, robot and manipulator planning to flaming actuation. 
This field is cleaned to provide a safe environment for the test. We have conducted 5 trials in total with random initial robot position in each trial. After the Spot robot arrives at the planned position, the RGB-D camera images are recorded to validate the detection-planning-actuation results. After the flaming has been completed, the thermal images are recorded for flaming coverage evaluation. Common weeds from southern Texas, such as common sunflower, giant ragweed, and smell melon, are identified and manually transplanted to the raised-bed plot for the experiments. 

\begin{figure}[!tbp]
    \centering
    \subfigure[]{\includegraphics[width = 1.05in]{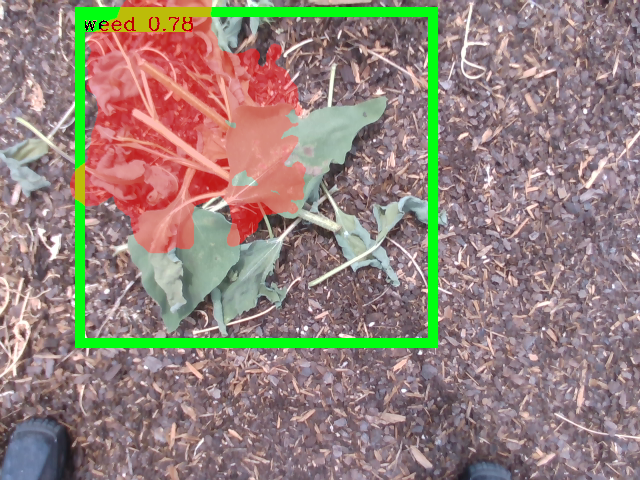}}
    \subfigure[]{\includegraphics[width = 1.05in]{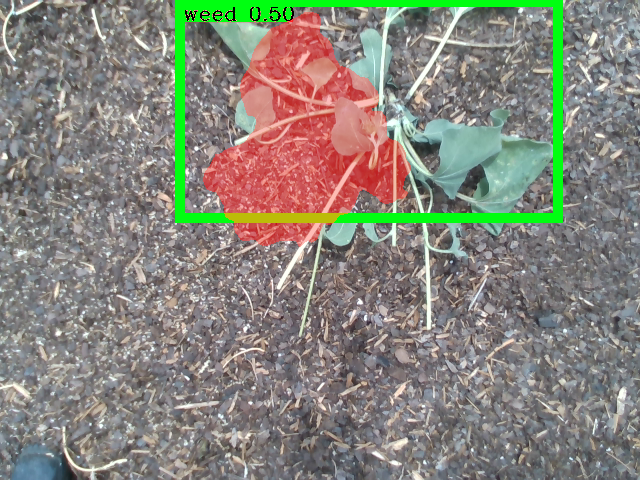}}
    \subfigure[]{\includegraphics[width = 1.05in]{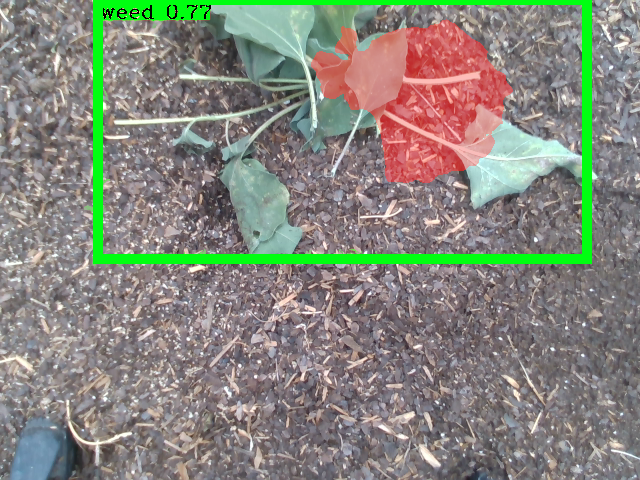}}
    \subfigure[]{\includegraphics[width = 1.05in]{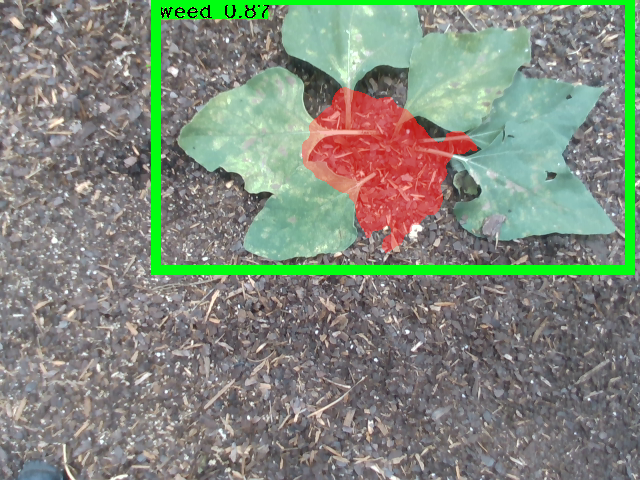}}
    \subfigure[]{\includegraphics[width = 1.05in]{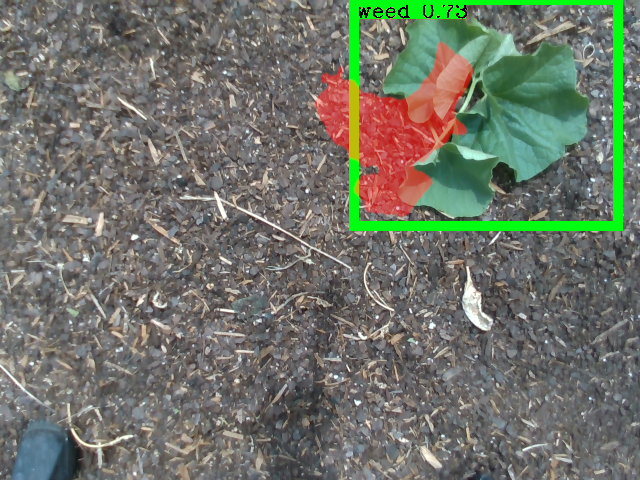}} \\
    \caption{Results of the rised-bed plot experiment trial 1 to 5. The green bounding boxes show the weed center detection results and the red areas are the re-projected hot spots captured by the thermal camera after flaming.}\label{fig:field-results}
\end{figure}

The results of the five raised-bed plot experiment trials are shown in Fig.~\ref{fig:field-results}. Due to the random initial positions, the weed positions in the images are different. The images from all five trials capture the weeds indicates that Spot detection-planning-actuation stage is successful and the robot has moved to region where the weed is reachable by the arm for the flaming action. After running the weed center detection and online flame estimation, the end-effector is controlled to approach the weed center with the flame tip based on its planned pose. The hot spots after flaming are recorded by the thermal camera and re-projected to the RGB images for evaluation. We measure the performance of the weed flaming system using the precision of the hot spot areas with respect to the weed center detection bounding boxes and the offset between the detected weed centers and the hot spot centers. Tab.~\ref{tab:field-test} shows the system precision and the flaming center offset in each trial and the result is satisfactory. The averaged precision indicates $94.4\%$ of the flammed area are within the detected weed bounding box, and the averaged center offset is $6.71$ cm.

\begin{table}[!htbp]
	\centering
	\caption{Rised-bed Plot Experiment Results}\label{tab:field-test}
  \begin{tabular}{c c c}
          \toprule[0.8pt]
           Trial Index &   Precision    &   Center Offset (cm)   \\\midrule
            1 & 96.7\% & 10.49 \\
            2 & 92.5\% & 5.72  \\
            3 & 100\% & 7.47 \\
            4 & 100\% & 2.54  \\
            5 & 82.6\% & 7.32 \\\midrule
            Avg. & 94.4\% & 6.71 \\
          \bottomrule
  \end{tabular}
\end{table}

The experiment in the cotton field aims to evaluate the performance of the weed flaming system in a real agriculture field. The experiment setup is shown in Fig.~\ref{fig:hardware_design}, and the results are included in the multimedia attachment. 

\section{Conclusions and Future Work}
We developed a robotic weed-flaming solution using a 6 DoF manipulator mounted on a quadrupedal robot. We presented the overall design, hardware integration, and software pipeline of the mobile manipulator system. We proposed a flame model with an estimation method that uses a center arc curve with a Gaussian cross-section model to describe the flame coverage in real time. The experiments have shown that our system and algorithm design have been successful.

In the future, we will plan the motion of the mobile manipulator to achieve dynamic flame coverage of multiple weeds. New weed removal techniques, such as electrocution, can also be integrated into the robot. New algorithms will be developed to enable the new system and improve efficiency.

\section*{}
{\small
\section*{Acknowledgment}
We are grateful to Aaron Kingery and Fengzhi Guo for their inputs and contributions.

\bibliographystyle{fmt/IEEEtran}
\bibliography{bib/di, bib/syxie, bib/Hu, bib/Joe}
}

\end{document}